\documentclass{article}
\usepackage{spconf,amsmath,graphicx}
\usepackage{xcolor}	
\usepackage{array,multirow}
\usepackage{booktabs}
\usepackage{comment}
\usepackage{hyperref}
\usepackage{amssymb}

\title{Choice Fusion as Knowledge for Zero-Shot Dialogue State Tracking}
\name{Ruolin Su$^1$ \ \ Jingfeng Yang$^2$ \ \ Ting-Wei Wu$^1$ \ \ Biing-Hwang Juang$^1$}
\address{$^1$ Georgia Institute of Technology, Atlanta, GA, USA\\ 
$^2$ Amazon, Palo Alto, CA, USA\\
ruolinsu@gatech.edu ~\ jingfengyangpku@gmail.com ~\ waynewu@gatech.edu ~\ juang@ece.gatech.edu}
\begin{document}
%
\maketitle
\begin{abstract}
With the demanding need for deploying dialogue systems in new domains with less cost, zero-shot dialogue state tracking (DST), which tracks user's requirements in task-oriented dialogues without training on desired domains, draws attention increasingly.
Although prior works have leveraged question-answering (QA) data to reduce the need for in-domain training in DST, they fail to explicitly model knowledge transfer and fusion for tracking dialogue states.
To address this issue, we propose CoFunDST, which is trained on domain-agnostic QA datasets and directly uses candidate choices of slot-values as knowledge for zero-shot dialogue-state generation, based on a T5 pre-trained language model.
Specifically, CoFunDST selects highly-relevant choices to the reference context and fuses them to initialize the decoder to constrain the model outputs.
Our experimental results show that our proposed model achieves outperformed joint goal accuracy compared to existing zero-shot DST approaches in most domains on the MultiWOZ 2.1.
Extensive analyses demonstrate the effectiveness of our proposed approach for improving zero-shot DST learning from QA.
\end{abstract}
\section{Introduction}
\label{sec:intro}

 Dialogue service automation nowadays has become an increasingly important part of modern commercial activities and calls for advanced dialogue systems to reduce the labor cost of human agents~\cite{pieraccini2009we}.
 Particularly, a task-oriented dialogue system aims to assist users in completing specific tasks, such as ticket booking, online purchasing, or travel planning, via a natural language dialogue~\cite{wen2016network}.
 Particularly, dialogue state tracking (DST) is the core function of a task-oriented dialogue system that tracks and generates dialogue states based on the user's requirements in a human-machine dialogue~\cite{bordes2016learning}.
 Typical dialogue states are collections of \textit{slot-value} pairs, for example, \textit{(destination, Cambridge)} and \textit{(day, Wednesday)} of a train ticket booking service.

  \begin{figure*}[htb]
  \centering
  \includegraphics[width=\linewidth]{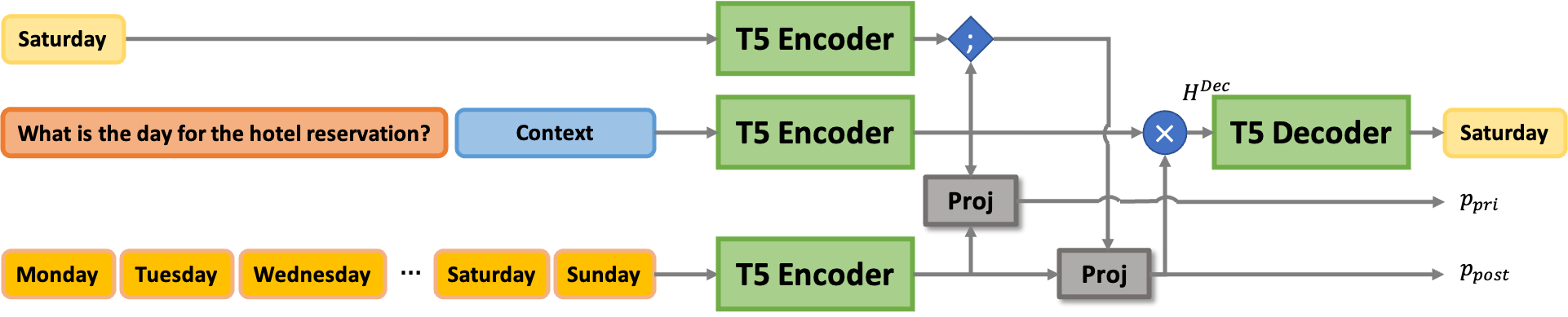}
  \caption{The overview of Choice-Fusion Dialogue State Tracking (CoFunDST). The input is a concatenated query consisting of a question (Tangerine) and reference context (Blue). A sequence of choice tokens (Orange) is used for training and reference. The golden answer token (Yellow) is an extra input during training as extra supervision for encoding the concatenated query. }
  \label{fig:t5}
\end{figure*}

 Nowadays, the requirements of deploying an increasing number of services across a variety of domains raise challenges to DST models in production~\cite{rastogi2020towards}. 
 However, existing dialogue datasets only span a few domains, making it impossible to train a DST model upon all conceivable conversation flows~\cite{campagna2020zero}.
 Furthermore, dialogue systems are required to infer dialogue states with dynamic techniques and offer diverse interfaces for different services.
 Despite the fact that the copy mechanism~\cite{Wu2019transferable} or dialogue acts~\cite{su2022act}
 are leveraged to efficiently track slots and values in the dialogue history,
 the performance of DST still relies on a large number of annotations of dialogue states, which is expensive and inefficient to collect data for every new domain and service.
 
 
 The lack of availability of annotated and representative data can limit the performance of DST at a large scale.
 A line of work suggests jointly encoding the schema and the dialogue context for DST to address the above challenge~\cite{ruan2020fine,lee2021dialogue}.
 On the other hand, based on the language processing theory~\cite{knoeferle2019predicting}---meanings in similar context could be understood and predicted before encountered, large-scale QA datasets provide an option to transfer learned knowledge for DST with little to no in-domain data without the loss of performance.~\cite{lin2021zero,li2021zero}.
 However, 
 none of the existing works model candidate choices in DST explicitly, resulting in a lack of efficiency and interpretability in using knowledge extraction capabilities from the QA datasets.
 
 
 As a means of tackling these issues, we propose \textbf{C}h\textbf{o}ice-\textbf{Fu}sio\textbf{n} \textbf{D}ialogue \textbf{S}tate \textbf{T}racking (\textbf{CoFunDST}) that trains on extensive QA datasets with sufficient annotations for zero-shot DST without training on particular domains\footnote{Code is publicly available at \url{https://github.com/youlandasu/Choice-Fusion}.}.
 Specifically, it fuses candidate choices as knowledge to predict slot-values accurately based on a T5~\cite{2020t5} pre-trained encoder-decoder language model.
 CoFunDST formulates DST and QA as machine reading comprehension (MRC)~\cite{hirschman1999deep,rajpurkar-etal-2016-squad} which generates the answer given the reference context.
 As a part of CoFunDST, we design appreciative choice selection to assess the relevance of all available candidate choices to the reference context and compute a probability distribution over these choices.
 Then we apply context-choice fusion~\cite{wu2020diverse} to incorporate the context-dependent choices as knowledge for initializing the decoder.
 Our work advances zero-shot DST in two ways: 1) for the first time, we model candidate choice of values as a distinctive resource of knowledge to leverage missing details for predicting slot-values accurately, 2) we propose the context-choice fusion to selectively incorporate encoded choices based on the dialogue context. 
 The performance of our model is demonstrated on the MultiWOZ 2.1~\cite{eric2019multiwoz} showing it outperforms existing zero-shot DST approaches in terms of joint goal accuracy in ``Restaurant'', ``Train'', and ``Taxi'' domains.
 Further analysis shows its effectiveness of the choice fusion and knowledge transfer from QA to DST generating different types of slot-values.

\begin{table*}
  \centering
  \begin{tabular}{lccccc}
    \toprule
    \multirow{2}{*}{Model} & \multicolumn{5}{c}{Joint Goal Accuracy} \\\cmidrule{2-6}
   & Hotel   & Train & Restaurant & Attraction   & Taxi  \\
    \midrule
    TRADE~\cite{Wu2019transferable} & 14.20 & 22.39 & 12.59 & 20.06  & 59.21  \\
    MA-DST~\cite{Kumar2020madst} & 16.28 & 22.76 & 13.56 & 22.46   & 59.27  \\
    SUMBT~\cite{lee-etal-2019-sumbt} & 19.80 & 22.50 & 16.50 & 22.60  & 59.50  \\
    TransferQA (T5-Small)~\cite{lin2021zero} & \textbf{21.82} & 25.66 & 17.98 & \textbf{26.14}  & 59.68  \\
    CoFunDST (T5-Small) & 21.07 & \textbf{25.95}  & \textbf{18.13} & 24.79  & \textbf{60.19} \\

    \bottomrule
  \end{tabular}
 \caption{Zero-shot joint goal accuracy on MultiWoz 2.1~\cite{eric2019multiwoz}. Model results of TRADE~\cite{Wu2019transferable}, MA-DST~\cite{Kumar2020madst}, and SUMBT~\cite{lee-etal-2019-sumbt} are obtained from reference papers. TransferQA (T5-Small) followes all setups in~\cite{lin2021zero}, but is trained using T5-Small. }
 \label{tab:zero-shot}
\end{table*}

\begin{table*}
  \centering
  \begin{tabular}{p{0.08\textwidth}p{0.15\textwidth}p{0.15\textwidth}p{0.15\textwidth}p{0.15\textwidth}p{0.15\textwidth}}
    \toprule
    \multirow{2}{*}{Settings} & \multicolumn{1}{c}{Hotel} & \multicolumn{1}{c}{Train} & \multicolumn{1}{c}{Restaurant} & \multicolumn{1}{c}{Attraction} & \multicolumn{1}{c}{Taxi} \\
    & JGA  ~/~ SGA ~/~ F1 & JGA  ~/~ SGA ~/~ F1 & JGA  ~/~ SGA ~/~ F1 & JGA  ~/~ SGA ~/~ F1 & JGA  ~/~ SGA ~/~ F1 \\
    \midrule
    KLD+Fuse & \textbf{21.04}~/~\textbf{67.13}~/~\textbf{39.74} & 23.84~/~\textbf{62.94}~/~\textbf{59.80} & \textbf{19.41}~/~\textbf{54.61}~/~\textbf{31.03} & \textbf{24.57}~/~\textbf{51.28}~/~\textbf{27.24} & 60.00~/~\textbf{75.56}~/~\textbf{68.78}\\
    KLD & 19.63~/~65.49~/~32.68 & \textbf{24.19}~/~60.91~/~52.04 & 16.55~/~54.71~/~29.54 & 18.00~/~47.04~/~19.54 & \textbf{60.06}~/~74.73~/~64.49 \\
    Fuse & 18.89~/~65.01~/~34.20 & 22.13~/~57.12~/~48.89 & 15.87~/~54.65~/~31.91 & 21.70~/~47.98~/~22.35 & 59.48~/~73.24~/~62.33 \\

    \bottomrule
  \end{tabular}
 \caption{Ablation study on the two components of the choice-fusion mechanism: appreciative choice selection with KLD loss (KLD) and context-choice fusion (Fuse). T5-Small~\cite{2020t5} is used and the evaluation results on joint goal accuracy (JGA), slot goal accuracy (SGA), and F1 in five domains of MultiWoz 2.1. }
 \label{tab:ablation}
\end{table*}

\section{Method}
\label{sec:method}
\subsection{Problem Formulation}
Both QA and DST are formulated as generative MRC problems, which take questions and choices as input and generate answers token-by-token by comprehending the reference context, as depicted in Fig.~\ref{fig:t5}.
For QA training, the input query combines the sequence of question tokens, \textit{i.e.} $q = \{q_1,q_2,\dots,q_K\}$, and the reference context tokens, \textit{i.e.} $c = \{c_1,c_2,\dots,c_L\}$, of the length $K$ and $L$, respectively. 
In other words, the model can be regarded as filling the sequence $a$ with correct tokens given the question $q$ and the reference context $c$: ``\textit{question: $q$ context: $c$ answer: [$a$]}'', where $a = \{a_1,a_2,\dots,a_M\}$ and $M$ is the length of answer.
Additionally, the concatenation of $N$ candidate choices of the given question is denoted as $v = \{v_1,v_2,\dots,v_N\}$.
We also encode the ground truth answer $\Tilde{a}$ in the training set into tokens to combine it with the encoded input. 

For DST reference, each domain slot is re-formulated as a natural language question in the form of ``\textit{What is the [slot] of [domain] that the user is interested in?}'', or ``\textit{What time}'' and ``\textit{How many}'' as prefixes of time- and number-related slots, similar to the domain-slot formulation in~\cite{lin2021zero}. In particular, the state value and dialogue context are taken as an answer and reference context, respectively.

\subsection{Choice-Fusion Mechanism} 
 \textbf{Appreciative Choice Selection:} 
 The appreciative choice selection is designed to select the choices that are highly relevant to the reference context, \textit{i.e.} the \textit{appreciative choices}.
 The choice tokens $v$ are processed by the T5 encoder as $V \in \mathbb{R}^{N\times T}$, where $T$ is the output hidden dimension of the Transformer.
 We calculate the \textit{prior} and \textit{posterior} probability distributions, $p_{pri}$ and $p_{post}$, of the candidate choices $V$ given the encoded question-context concatenation $D_{pri} \in \mathbb{R}^{(K+L)\times T}$ and the encoded golden answer $D_{post} \in \mathbb{R}^{M\times T}$ . 
 Note that only the prior distribution of Eq.~\ref{eq:1} is used during reference.
\begin{equation}
\label{eq:1}
\begin{aligned}
p_{pri} &= \text{softmax} (\tanh(V W^V)  \tanh([D_{pri}] {W^D}_{pri}))\\
\end{aligned}
\end{equation}
\begin{equation}
\label{eq:2}
\begin{aligned}
p_{post} &=\text{softmax} (\tanh(V W^V)  \tanh([D_{pri} ; D_{post}] {W^D}_{post}))\\
\end{aligned}
\end{equation}
 where $W^{D}_{pri} \in \mathbb{R}^{T\times F}$, ${W^{D}}_{post}\in \mathbb{R}^{2T \times F}$, and $W^V \in \mathbb{R}^{T \times F}$ are trainable parameter matrices, $F$ is the intermediate dimension for projecting context on the choices.
 Then the objective function is the Kullback–Leibler divergence (KLD)~\cite{kullback1951information} to optimize the distance of \textit{prior} and \textit{posterior} distributions of $V$, where the ground truth answer embedding $D_{post}$ is served as the posterior knowledge for choice selection (Eq.~\ref{eq:kld}). 
\begin{equation}
\label{eq:kld}
    \mathcal{L}_{KLD} =KLD(p_{pri},p_{post})
\end{equation}

 \textbf{Context-Choice Fusion:} 
 The context-choice fusion leverages appreciative choices to address choice knowledge in answer generation, by fusing the context and appreciative choices to the decoder.
 To fuse the obtained appreciation over choices for generating accurate answers, candidate choices are weighted by the posterior distribution $p_{post}$  to initialize the input of the decoder, as shown in Eq.~\ref{eq:decoder-init}.
\begin{equation}
\begin{aligned}
H^{Dec} &= \tanh([D_{pri} ; p^{T}_{post}V]\cdot W^{Dec}))\\
\end{aligned}
\label{eq:decoder-init}
\end{equation}
where $H^{Dec} \in 	\mathbb{R}^{(K+L+N)\times T}$ is the fused input to the decoder, and $W^{Dec} \in \mathbb{R}^{T\times T}$. 
For inference, the prior distribution $p_{pri}$ is used in replace of $p_{post}$ for Eq.~\ref{eq:decoder-init}.
Such that the appreciative choices are contextualized and incorporated into the CoFunDST model as knowledge.

The overall objective in Eq.~\ref{eq:loss} is the sum of the KLD and the Cross-Entropy loss of the decoder output and the ground truth answer $\Tilde{a}$, where non-categorical slots and categorical slots are jointly trained. 
\begin{equation}
\label{eq:loss}
\mathcal{L} = -\log P(a=\Tilde{a}|D_{pri}, D_{post},V) + \mathcal{L}_{KLD}
\end{equation}

\section{Experiments} 
\label{sec:exp}

\textbf{Datasets:} The model is pre-trained on 20\% of the combination of six extractive QA datasets
and two multi-choice QA datasets following the dataset pre-processing and slicing in~\cite{lin2021zero}.
To verify the generalization among different domains, we evaluate models on MultiWOZ 2.1~\cite{eric2019multiwoz} and follow dataset setups in~\cite{Wu2019transferable}.
There are 30 distinguished domain-slots in MultiWOZ 2.1 in total, where there are 12 categorical slots provided with collections of values and 18 non-categorical slots.

\textbf{Baselines:} We select the following models as zero-shot DST baselines. (1) TRADE~\cite{Wu2019transferable} is an encoder-decoder model which leverages slot gates and copy mechanism and shares parameters for predicting unseen slot-values. (2) SUMBT~\cite{lee-etal-2019-sumbt} uses pre-trained BERT~\cite{devlin2018bert} to learn the relations between slot types and values appearing in utterances and predict dialogue states with slot-utterance matching. (3) MA-DST~\cite{Kumar2020madst} encodes dialogue context and domain-slots with attention mechanisms at multiple granularities to learn at different semantic levels. (4) TransferQA~\cite{lin2021zero} proposes a task-transfer framework and takes the combination of slot, values, and dialogue context as the input for zero-shot DST.

TRADE~\cite{Wu2019transferable}, MA-DST~\cite{Kumar2020madst}, and SUMBT~\cite{lee-etal-2019-sumbt} are evaluated in the cross-domain setting, where the models are trained on the four domains in MultiWOZ 2.1 and evaluated on the held-out domain. 
On the contrary, TransferQA~\cite{lin2021zero} and our model are trained on the combined QA dataset only. Therefore, it is unnecessary to use in-domain DST data.
Although TransferQA and ours are both based on domain-agnostic QA training, we extend this idea with knowledge fusion and directly use values in domain ontology as candidate choices to align predictions of DST with the training procedure.

\textbf{Implementation:} 
We implemented our model based on T5-Small~\cite{2020t5}, which is a pre-trained encoder-decoder model for natural language generation. The input text of the T5 encoder is truncated to 512 tokens. 
The intermediate dimension for prior and posterior probabilities is 64. The fused input to the T5 decoder is passed through a 512-unit Feed-Forward layer for initialization.
Adafactor~\cite{shazeer2018adafactor} is used as the optimizer with initial learning rate 3e-4 and warm-up steps 100. The number of training epochs is 6.
As for our implementation of TransferQA~\cite{lin2021zero}, all other hyper-parameters are kept the same as the original settings except for the model size. 


\section{Results}
The zero-shot joint goal accuracy (\textbf{JGA}), which is the average accuracy of predicting all slot-values of a turn correctly, on MultiWOZ 2.1 is presented in Table~\ref{tab:zero-shot}. 
It can be seen that our model exceeds all baselines on JGA in ``\textit{restaurant}'', ``\textit{taxi}'', and ''\textit{train}`` domains with KLD loss and context-choice fusion adopted.
Compared to TransferQA, our model does not combine candidate choice tokens with the input question and context which tend to be truncated due to the lengthy inputs. 
Furthermore, CoFunDST maintains context information by fusing the contextualized weights of candidate choices for inference.
Experimental results demonstrate our proposed method effectively generalize from QA to new domains without annotated data for DST.
The JGA is somewhat less in the other two domains where multiple-choice slots mostly contain an extensive collection of choices to be efficiently used for context-choice fusion.
For example, the average number of choices for categorical slots in the domain ontology of ``\textit{restaurant}'' is 5, while that of ``\textit{attraction}'' is 11.
It indicates that more choices associated with a slot lead to less efficient incorporation of choice knowledge into the model.
Compared to the first three rows in Table 1 based on leave-one-out training, our model outperformed all of them, indicating it is unnecessary to perform in-domain training.

\begin{table}[t!]
  \centering
  \resizebox{\linewidth}{!}{
  \begin{tabular}{lccccc}
    \toprule
    \textbf{Domain} & \textit{\#non-cat} & \textit{\#cat} & \textit{Non-Categorical} & \textit{Categorical} & \textit{All}\\
    \midrule
    Hotel & 4911&7609 & 66.91 & 55.83 & 68.05 \\
    Train & 8832&2514 & 61.62 & 39.24 & 68.12\\
    Restaurant & 5800&4967 & 60.14 & 45.69 & 62.27\\
    Attraction & 1249&3253 & 51.61 & 35.38 & 52.94\\
    Taxi & 1753&0 & 75.55  & - & 75.55\\

    \bottomrule
  \end{tabular}
  }
 \caption{The slot goal accuracy of non-categorical and categorical slots on MultiWoz 2.1, respectively. \textit{\#non-cat} and \textit{\#cat} are the total numbers of non-categorical and categorical slots for dev.}
 \label{tab:ex_mc}
\end{table}

\section{Discussion}
\label{sec:dis}

We evaluate two components in the choice-fusion mechanism, \textit{i.e.} the appreciative choice selection with KLD loss (KLD) and the context-choice fusion (Fuse). 
Table~\ref{tab:ablation} summarizes the results of using our proposed techniques, where the slot goal accuracy (\textbf{SGA}) is the average accuracy of predicting the value of a slot correctly and F1 is the harmonic mean of the precision and recall to evaluate the performance of slot-value predictions.
It is evident that the model adopting both modules essentially outperforms the models with only one module.
The SGA and F1 drop when not using the context-fused result for the decoder, indicating that the context fusion is important for constraining the generation of slot-values at the slot level. 
It is noted that JGA drops for all except ``train'' and ``restaurant'' domains slightly compared to applying KLD loss only.
This suggests that the turn-level generation of dialogue states could be impacted by re-initializing the decoder, while slot-level accuracy and F1 still outperform. 
Moreover, it is observed that in all cases metrics drop significantly without KLD loss, which proves that the improved alignment of the prior and posterior distributions of choices depending on the context will benefit accurate value generation.
Empirically, it seems that leveraging ground-truth annotations is better than only fusing choices dependent on the prior context in training because the appreciative choice selection is superior to context-choice fusion on all metrics.

The SGA by domains and slot types in Table~\ref{tab:ex_mc} shows that our model outperforms three more challenging domains with more non-categorical slots than categorical ones.
Non-categorical slots tend to have a larger vocabulary size of values and thus benefit from joint training using knowledge fusion.
The higher SGA for predicting non-categorical slots is probably owing to more extractive data used in QA training, which results in the model bias that is more likely to generate extractive values.
As our focus in this paper is on choice selection and fusion, we leave the study on related datasets as future work.

To explain the performance in the three domains we outperform, we randomly sample 50 dialogues in predictions for analyses. We note that the average number of candidate choices  of categorical slots per dialogue in the ``Restaurant'', ``Train'', and ``Taxi'' domains are mostly far less than that in the ``Attraction'' and ``Hotel'' domains, except for ``Train'' and ``Hotel'' domains that are 7 and 5.7, respectively. However, the ``Hotel'' domain consists of 6 different categorical slots while there is only one categorical slot in the ``Train'' domain. Further error analysis in “Attraction” and “Hotel” which are degraded shows 87.09\% of slot errors are missing to predict a NOT none (active) slot in the “Attraction” domain and 72.22\% of them are from categorical slots. While there are 72.06\% missing errors in the “Hotel” domain and 70.8\% of them are from categorical slots. These suggest the degradation in the two domains is mostly due to not predicting all related categorical slots as active, and this is more likely to happen when the number of categorical slots is many in the “Attraction” and “Hotel” domains.

\section{Conclusion}

We introduce the CoFunDST model that incorporates candidate choices and transfers knowledge from QA for zero-shot dialogue state generation. 
The appreciative choice-selection module selects candidate choices that are highly relevant to the reference context.
And the context-choice fusion module uses the context and the appreciative choices to initialize the decoder.
Our model achieves outperformed zero-shot joint goal accuracy on multiple domains of MultiWOZ 2.1 and the choice-fusion mechanism is shown effective for improving domain-agnostic training for generalization to new domains.
Incorporating other domain-agnostic knowledge for zero-shot dialogue state tracking is the area of study that we wish to extend and develop in the future.



\bibliographystyle{IEEEbib}
\bibliography{strings,refs}

\end{document}